\documentclass[preprint,12pt]{elsarticle}

\usepackage{graphicx}
\usepackage{amsmath}
\usepackage{amssymb}
\usepackage{algorithmic}
\usepackage{algorithm}
\usepackage{subfigure}
\usepackage{array}

\newcommand{\bfx}{{\textbf{x}}}

\newcommand{\bfw}{{\textbf{w}}}

\journal{Neurocomputing}

\begin{document}

\begin{frontmatter}

\title{Semi-supervised learning of local structured output predictors}

\author[X]{Xin Du}
\ead{xindu.njtu@gmail.com}

\address[X]{School of Electrical Engineering, Beijing Jiaotong University, Beijing 100044, China}

\begin{abstract}
In this paper, we study the problem of semi-supervised structured output prediction, which aims to learn predictors for structured outputs, such as sequences, tree nodes, vectors, etc., from a set of data points of both input-output pairs and single inputs without outputs. The traditional methods to solve this problem usually learns one single predictor for all the data points, and ignores the variety of the different data points. Different parts of the data set may have different local distributions, and requires different optimal local predictors. To overcome this disadvantage of existing methods, we propose to learn different local predictors for neighborhoods of different data points, and the missing structured outputs simultaneously. In the neighborhood of each data point, we proposed to learn a linear predictor by minimizing both the complexity of the predictor and the upper bound of the structured prediction loss. The minimization is conducted by gradient descent algorithms. Experiments over four benchmark data sets, including DDSM mammography medical images, SUN natural image data set, Cora research paper data set, and Spanish news wire article sentence data set, show the advantages of the proposed method.
\end{abstract}


\end{frontmatter}

\section{Introduction}
\label{sec:Introduction}

Machine learning refers to the problem of learning a predictive model to predict a output from a input data point \cite{Fong20151725,Wang20151693,fan2011margin,fan2015improved,Wang2015125,Singh20158609,Li2015,Xu2015,wang2015multiple,wang2014effective,liu2015supervised,lin2016multi}. The forms of output are various, usually including binary class label and continues response. The problem of predicting binary class label is called classification \cite{Shu2015230,fan2010enhanced,fan2014finding,fan2014tightening,chen2015towards,fu2015bayesian,wang2015supervised,wang2015representing,zhou2014biomarker}, while the problem of predicting continues response is called regression \cite{Kavitha2015478,PascualGonzlez201534,Yang201540,PiaMonarrez2015244}. Both of these two problems have many applications, such as computer vision, natural language processing, bioinformatics, and finance. However, in many these applications, the forms of outputs of the prediction may be beyond binary class labels and continues responses. For example, in the part-of-speech tagging problem of natural language processing, given a sequence of words, we want to predict the tags of the part-of-speech of the works, and the output of the prediction is a sequence of parts-of-speech \cite{Vincze201544,Fonseca2015,Bhowmik2015,Carneiro201511,Wang2015458}. In the problem of hierarchical image classification problem, the class labels of images are organized as a tree structure, and the outputs of the prediction problem are the leaves of a tree \cite{Baravalle2015,Gao2015,Decost2015126,Gao201548}. In this case, the predictive models designed for binary class labels and continues responses cannot handle these output forms, and new predictive model should be developed. The output forms other than binary labels and continues responses are called structured outputs. The structured outputs include a wide range of types of outputs, such as sequences, vectors, graph nodes, tree leaves, etc. The problem of learning predictive models to predict unknown structured outputs are called as structured output prediction \cite{Srikumar20143266,Jiang2014i609,Han20141665,NouraniVatani2015173,Jiang2015}. Given a set of input-output data points, where the outputs are structured, this problem usually learns a predictive model to match the input-output relationship. Most existing methods designed to solve assumes that in the training set, all the input data points have their corresponding outputs. However, in real-world application, many output are not available for the inputs \cite{altun2005maximum,brefeld2006semi,suzuki2007semi,jiang2015manifold}. The training set is composed of two parts. One part is a set of input-output pairs, which is called labeled set. The other part is a set of single input data points with missing outputs, and this set is called unlabeled data set. Learning from such a training set is call semi-supervised learning \cite{gan2016towards,zeng2016semi,wang2016enhancing}. In this paper, we invest the problem of learning structured output predictors from such a training set. This problem is called semi-supervise structured output prediction.

\subsection{Related works}

Our work is a novel semi-supervised structured output prediction method, thus we introduce the related works of this direction. There are a number of existing algorithms for this problem, which are briefly introduced as follows.

\begin{enumerate}
\item Altun et al. \cite{altun2005maximum} proposed the problem of semi-supervised learning with structured outputs. Moreover, a novel discriminative approach was also proposed to use the manifold of input features of both labeled and unlabeled data points. This approach is based on the semi-supervised maximum-margin formulation. It is an inductive algorithm, and it can be easily extended to new-coming test data points.

\item Brefeld and Scheffer \cite{brefeld2006semi} proposed to solve the problem of semi-supervised structured output prediction by learning in the space of input-output space, and using co-training method. This method is based on the assumption that the multiple structured output predictors should be consent with each other. Based on this assumption, the structural support vector machine is extended to the argued input-output space.

\item Suzuki et al. \cite{suzuki2007semi} proposed a hybrid method to solve the problem of semi-supervised structured output learning. This method combines both the generative and discriminative methods. The objective of this method is composed of log-linear forms of both discriminative structured predictor and generative model. The generative model is used to incorporate unlabeled data points. The discriminant functions is enhanced by the unlabeled data points provided by the generative model.

\item Jiang et al. \cite{jiang2015manifold} proposed to regularize the structured outputs by the manifold constructed from the input space directly. This method constructs a nearest neighbor graph from the input features, and use it to represent the manifold. Then the manifold is used to regularize the learning of the missing outputs of the unlabeled data points. The outputs and the predictor are learned simultaneously, and they regularize each other in the learning process.

\end{enumerate}

{Our work approximate the upper bound of the structured loss, and is inspired by the lower bound approximation of the structure learning of the Bayesian network \cite{fan2014tightening,fan2015improved}. Thus we also discuss the works of bound approximation technologies of \cite{fan2014tightening,fan2015improved}.
\begin{enumerate}
\item Fan et al. \cite{fan2014tightening} proposed to tighten the upper and lower bounds of the breadth-first branch and bound algorithm for the learning of Bayesian network structures. The informed variable groupings is used to create the pattern databases to tighten the lower bounds, while the anytime learning algorithm is used to tighten the upper bound. These strategies show good performance in the learning process of the Bayesian network structures. The work of \cite{fan2014tightening} is a contribution of major significance to the bound approximation community, and our upper bound approximation method is also based on these strategies.
\item Fan et al. \cite{fan2015improved} further proposed to improve the lower bound function of static k-cycle conflict heuristic for the learning of Bayesian network structures. This work is used to guild the search of the most promising search spaces. It use a partition of the random variables of a data set, and the further research is based on the importance of the partition. A new partition method was proposed, and it uses the information extracted from the potentially optimal parent sets.
\end{enumerate}}

\subsection{Our contributions}

All the mentioned semi-supervised structured output prediction methods learns one single predictor for the entire data set. However, we observe that a training set, the local distributions of neighborhoods play important roles in the problem of modeling of both input and structured outputs. It is extremely important to respect the local distributions when the structured output predictor are learned. This is even more important for learning from semi-supervised data sets. This is because for this type of data set, only a few data points have available structured outputs, and the structured outputs of all other data points are missing. To learn the missing structured outputs, we need to explore the connections between different data points, so that we may propagate the structured outputs from the labeled set to the unlabeled set. It has been shown that using local connections is an effective way to model the connections among different data points  \cite{jiang2015manifold}. To explore the local distributions, one option is to construct a nearest neighbor graph, and use it to regularize the learning of the predictors. More specifically, with the nearest neighbor graph, we hope that the neighboring data points can obtain similar structured outputs from the predictor \cite{hu2010manifold,jiang2015manifold}. However, one single predictor is usually not enough to characterize multiple local distributions, thus even we use the neighborhood graph to regularize the learning the predictor, it is still not guaranteed that the local distributions are sufficiently modeled with regard to the structured output prediction problem. This is an even more serious problem when it is applied to a semi-supervised data set. With such a data set, only a few data points have corresponding structured outputs, while most of the data points do not. The learned predictor can easily fits to the labeled data points.

To solve this problem, we propose to learn multiple local linear structured output predictor for different neighborhoods to model the local distributions, instead of learning one single predictor for the entire data \cite{zhang2012semi,xue2009local}. Moreover, we also propose to learn the missing structured outputs for a semi-supervised data set simultaneously. For each data point, we propose to present the local distribution around this data point by its $k$ nearest neighborhood, and model it by learning a local linear structured output predictor. To learn the parameters of this local predictor, we propose to minimize a upper bound of the structured losses of the data points in this neighborhood, and the squared $\ell_2$ norm of the predictor parameter vector. In this process, the predicted structured outputs are compared to the learned structured outputs. The learning of the structured outputs are simultaneously by the true structured outputs of the labeled data points, and the local predictors. Some data points are shared by different neighborhoods, and they play the role of bridging different local distributions to learn a complete manifold. To solve the problem, we develop an iterative algorithm, by using gradient descent method.

\subsection{Paper organization}

The rest parts of this paper are organized as follows. In section \ref{sec:Problem}, we model the learning problem, and present the optimization methods for the problem. In section \ref{sec:algorithm}, the iterative algorithm for the learning process, and the algorithm for the test process are both introduced. In section \ref{sec:Exp}, the experiments in four benchmark data sets are presented, including DDSM Mammography medical image data set, SUN natural image data set, Cora research paper data set, and Spanish news wire article sentence data set. In section \ref{sec:Conclu}, the paper is concluded. In section \ref{sec:futurework}, we discuss the future works.

\section{Problem modeling and optimization}
\label{sec:Problem}

\subsection{Problem modeling}

Suppose we have a training set of $n$ data points, $\mathcal{X} = \mathcal{L}\cup \mathcal{U}$, which is composed of a labeled subset, $\mathcal{L}$, and a unlabeled subset, $\mathcal{U}$. $\mathcal{L}$ contains $l$ data points of input-output pairs, $\mathcal{L} =\{(\bfx_i,\overline{y}_i)\}_{i=1}^l$, where $\bfx_i\in \mathbb{R}^d$ is the $d$-th dimensional feature vector input of the $i$-th data point, $\overline{y}_i\in \mathcal{Y}$ is the true structured output of the $i$-th data point, $\mathcal{Y}$ is the space of structured outputs. $\mathcal{U} $ contains $u = n-l$ data points of inputs without outputs, $\mathcal{U} = \{\bfx_i\}_{i=l+1}^n$. The structured output prediction problem is to learn predictors to predict the structured outputs from the inputs from the training set. We proposed to learn a local predictor for the neighborhood of each data points, instead of learning one single predictor for all the data points. We present the neighborhood of the $i$-th data point as the set of its $k$ nearest neighbors,

\begin{equation}
\label{equ:Ni}
\begin{aligned}
\mathcal{N}_i = \{\bfx_j:\bfx_j~is~the~l-th~nearest~neighbor~of~\bfx_i,~and~l\leq ~k \}.
\end{aligned}
\end{equation}
Given a data point of the $i$-th neighborhood, $\bfx_j\in \mathcal{N}_i$, to predict is structured output, we match it against all the possible structured outputs. Given a candidate structured output, $y\in \mathcal{Y}$, and an input feature vector, $\bfx_j$, we use a joint representation to match them. The joint representation of $y$ and $\bfx_j$ is denoted as $\Phi(\bfx_j,y) \in \mathbb{R}^m$, where $m$ is the dimension of the joint representation. We design a local linear predictor for the joint representations of the data points of the neighborhood of each data point, $\mathcal{N}_i$, and use it to predict the structured outputs,

\begin{equation}
\label{equ:llr}
\begin{aligned}
y^*_j = \underset{y\in \mathcal{Y}}{\arg\max}~ \bfw_i^\top \Phi(\bfx_j,y), ~j:\bfx_j\in \mathcal{N}_i,
\end{aligned}
\end{equation}
where $\bfw_i$ is the parameter of the local predictor of the $i$-th neighborhood, $\mathcal{N}_i$, and $y^*_j$ is the predicted structured output of a data point in $\mathcal{N}_i$, $\bfx_j$. Apparently, this local predictor match the joint representations of a input vector and each candidate output by a linear function, and return the output candidate which gives the maximum matching response.

Because the training set is composed of labeled and unlabeled subsets, the outputs of the data points in $\mathcal{U}$ are missing, we also proposed to learn the complete outputs simultaneously. The learned output set is denoted as $\{y_i\}|_{i=1}^n$, where $y_i$ is the outputs of the $i$-th data point. To guarantee the learned outputs are consistent with the given true outputs, we impose that the outputs of the data points in $\mathcal{L}$ is equal to the true outputs,

\begin{equation}
\label{equ:constraint}
\begin{aligned}
y_i = \overline{y}_i, i:(\bfx_i,\overline{y}_i) \in \mathcal{L}.
\end{aligned}
\end{equation}

We propose to learn the parameters of the local predictors, $\bfw_i|_{i=1}^n$, and the complete outputs, $y_i|_{i=1}^n$ simultaneously. To this end, we decompose the learning problem to each neighborhood. In the $i$-th neighborhood, we learn the local predictor parameter, $\bfw_i$, and the outputs of the data points in $\mathcal{N}_i$, $y_j|_{j:\bfx_j\in \mathcal{N}_i}$. We use a structured loss function to measure the loss of predicting the $j$-th structured output as $y_j^*$, while the corresponding output is $y_j$. The loss function is denoted as $\Delta(y_j, y_j^*)$. Moreover, we also propose to keep the predictor as simple as possible, by minimizing the squared $\ell_2$ norm of the local predictor, $\bfw_i$. By minimizing this loss function with regard to $\bfw_i$ and $y_j$ over the $i$-th neighborhood, and minimizing squared $\ell_2$ norm of $\bfw_i$, we obtain the optimal local predictor parameter,

\begin{equation}
\label{equ:localproblem}
\begin{aligned}
\min_{\bfw_i, y_j|_{j:\bfx_j\in \mathcal{N}_i}}~ &\left \{ \frac{1}{k} \sum_{j:\bfx_j\in \mathcal{N}_i} \Delta(y_j, y_j^*) + \frac{C}{2} \|\bfw_i\|_2^2\right \},\\
s.t.~&
y_j = \overline{y}_j, j:(\bfx_j,\overline{y}_i)\in \mathcal{L},
\end{aligned}
\end{equation}
where $C$ is a tradeoff parameter to balance the weights of the first and second terms of the objective.

Because the structured loss function is usually complex and difficult to optimize directly, we seek its upper bound and minimize it instead. According to (\ref{equ:llr}), we have,

\begin{equation}
\label{equ:upperbound}
\begin{aligned}
&\bfw_i^\top \Phi(\bfx_j,y^*_j) \geq \bfw_i^\top \Phi(\bfx_j,y_j), ~j:\bfx_j\in \mathcal{N}_i,\\
&\Rightarrow
\bfw_i^\top \left ( \Phi(\bfx_j,y^*_j) - \Phi(\bfx_j,y_j) \right )\geq 0,\\
&\Rightarrow
\bfw_i^\top \left ( \Phi(\bfx_j,y^*_j) - \Phi(\bfx_j,y_j) \right ) + \Delta(y_j, y_j^*) \geq \Delta(y_j, y_j^*),\\
&\Rightarrow
\max_{y_j'\in \mathcal{Y}}\left [\bfw_i^\top \left ( \Phi(\bfx_j,y'_j) - \Phi(\bfx_j,y_j) \right ) + \Delta(y_j, y_j')\right ] \geq \Delta(y_j, y_j^*).
\end{aligned}
\end{equation}
Thus the upper bound of $\Delta(y_j, y_j^*)$ can be given as

\begin{equation}
\label{equ:upperbound1}
\begin{aligned}
&\max_{y_j'\in \mathcal{Y}}\left [\bfw_i^\top \left ( \Phi(\bfx_j,y'_j) - \Phi(\bfx_j,y_j) \right ) + \Delta(y_j, y_j')\right ] \\
&= \left [\bfw_i^\top \left ( \Phi(\bfx_j,z^*_j) - \Phi(\bfx_j,y_j) \right ) + \Delta(y_j, z^*_j)\right ],
\end{aligned}
\end{equation}
where $z^*_j$ is defined as

\begin{equation}
\label{equ:upperbound2}
\begin{aligned}
&z^*_{i,j} = \underset{y_j'\in \mathcal{Y}}{\arg\max}
\left [\bfw_i^\top \left ( \Phi(\bfx_j,y'_j) - \Phi(\bfx_j,y_j) \right ) + \Delta(y_j, y_j')\right ],
\end{aligned}
\end{equation}
and it is an important parameter to define the upper bound of the loss function. With the upper bound, the minimization problem can be transferred to the following problem,

\begin{equation}
\label{equ:localproblem1}
\begin{aligned}
\min_{\bfw_i, y_j|_{j:\bfx_j\in \mathcal{N}_i}}~ &\left \{ \frac{1}{k} \sum_{j:\bfx_j\in \mathcal{N}_i}
\left [\bfw_i^\top \left ( \Phi(\bfx_j,z^*_{i,j}) - \Phi(\bfx_j,y_j) \right ) + \Delta(y_j, z^*_{i,j})\right ]
+ \frac{C}{2} \|\bfw_i\|_2^2\right \}\\
s.t.~&
y_j = \overline{y}_j,j:(\bfx_j,\overline{y}_i)\in \mathcal{L}.
\end{aligned}
\end{equation}

To bridge the learning problems of different neighborhoods, we propose to combine them into one single problems over the entire data set,

\begin{equation}
\label{equ:localproblem2}
\begin{aligned}
\min_{(\bfw_i, y_i)|_{i=1}^n}~ &\sum_{i=1}^n\left \{ \frac{1}{k} \sum_{j:\bfx_j\in \mathcal{N}_i}
\left [\bfw_i^\top \left ( \Phi(\bfx_j,z^*_{i,j}) - \Phi(\bfx_j,y_j) \right ) + \Delta(y_j, z^*_{i,j})\right ]
+ \frac{C}{2} \|\bfw_i\|_2^2\right \}\\
s.t.~&
y_i = \overline{y}_i,~i:(\bfx_i,\overline{y}_i)\in \mathcal{L}.
\end{aligned}
\end{equation}
The motive to combine the local learning problems to one single over all problem is to connect these local problems by the overlapping data points of different neighborhoods. For example, $\mathcal{N}_i$ and $\mathcal{N}_{i'}$ have overlapping data points, and one of them is $\bfx_j$, i.e., $\bfx_j\in \mathcal{N}_i$, and $\bfx_j\in \mathcal{N}_{i'}$. Then the learning of $y_i$ will be a common process of both the local problems of $\mathcal{N}_{i}$ and $\mathcal{N}_{i'}$, and it can also regularize the learning of both $\bfw_i$ and $\bfw_{i'}$.

\subsection{Problem optimization}

To solve the problem in (\ref{equ:localproblem2}), we propose to use an iterative algorithm. In this algorithm, we use an alternate optimization strategy. Each iteration has two steps. In the first step, we fix $y_i|_{i=1}^n$ and update $\bfw_i|_{i=1}^n$ by gradient descent algorithm. In the first step, we fix the updated $\bfw_i|_{i=1}^n$ and update $y_i|_{i=1}^n$ one by one.

\subsubsection{Updating $\bfw_i|_{i=1}^n$}

When the outputs $y_i|_{i=1}^n$ are fixing, the problem in (\ref{equ:localproblem2}) is transferred to

\begin{equation}
\label{equ:localproblem3}
\begin{aligned}
\min_{\bfw_i|_{i=1}^n}~ &\sum_{i=1}^n\left \{ \frac{1}{k} \sum_{j:\bfx_j\in \mathcal{N}_i}
\left [\bfw_i^\top \left ( \Phi(\bfx_j,z^*_{i,j}) - \Phi(\bfx_j,y_j) \right ) + \Delta(y_j, z^*_{i,j})\right ] \right. \\
&\left.
\vphantom{
\left \{ \frac{1}{k} \sum_{j:\bfx_j\in \mathcal{N}_i}
\left [\bfw_i^\top \left ( \Phi(\bfx_j,z^*_{i,j}) - \Phi(\bfx_j,y_j) \right ) + \Delta(y_j, z^*_{i,j})\right ] \right.
}
+ \frac{C}{2} \|\bfw_i\|_2^2 = g(\bfw_i)\right \},
\end{aligned}
\end{equation}
where $g(\bfw_i)$ is the objective of the problem of the $i$-th neighborhood. To update $\bfw_i$, we use the gradient descent algorithm. However, in the objective, $z_j^*$ is also a function of $\bfw_i$, thus it is difficult to obtain the gradient function directly. To solve this problem, instead of seeking the gradient function, we seek the sub-gradient function of $g(\bfw_i)$ with regard to $\bfw_i$. We first update each $z_j^*$ according to previously updated $\bfw_i$, and then fix it to obtain the sub-gradient function with regard to $\bfw_i$,

\begin{equation}
\label{equ:localproblem4}
\begin{aligned}
\nabla g(\bfw_i) =  \frac{1}{k} \sum_{j:\bfx_j\in \mathcal{N}_i}
\left [ \left ( \Phi(\bfx_j,z^*_{i,j}) - \Phi(\bfx_j,y_j) \right ) \right ]
+ C\bfw_i.
\end{aligned}
\end{equation}
With the sub-gradient function, the updating rule of $\bfw_i$ is given as follows,

\begin{equation}
\label{equ:updatewi}
\begin{aligned}
\bfw_i &\leftarrow \bfw_i - \eta \nabla g(\bfw_i)\\
& = \bfw_i - \eta\left ( \frac{1}{k} \sum_{j:\bfx_j\in \mathcal{N}_i}
\left [ \left ( \Phi(\bfx_j,z^*_{i,j}) - \Phi(\bfx_j,y_j) \right ) \right ]
+ C\bfw_i \right )\\
& = (1-\eta C)\bfw_i +  \frac{\eta}{k} \sum_{j:\bfx_j\in \mathcal{N}_i}
\left [ \left (  \Phi(\bfx_j,y_j)- \Phi(\bfx_j,z^*_{i,j}) \right ) \right ].
\end{aligned}
\end{equation}
We can see that the updating function of $\bfw_i$ is a combination of a weighted $\bfw_i$ and a function of the joint representations of the data points in $\mathcal{N}_i$.

\subsubsection{Updating $y_i|_{i=1}^n$}

When $\bfw_i|_{i=1}^n$ are fixed and only $y_i|_{i=1}^n$ are considered, we remove the terms irrelevant to $y_i|_{i=1}^n$ in problem of (\ref{equ:localproblem2}), and transfer it to the following problem,

\begin{equation}
\label{equ:localproblem5}
\begin{aligned}
\min_{y_i|_{i=1}^n}~ &\sum_{i=1}^n\left \{ \frac{1}{k} \sum_{j:\bfx_j\in \mathcal{N}_i}
\left [\Delta(y_j, z^*_{i,j}) - \bfw_i^\top \Phi(\bfx_j,y_j) \right ] \right \}\\
s.t.~&
y_i = \overline{y}_i,~i:(\bfx_j,\overline{y}_i)\in \mathcal{L}.
\end{aligned}
\end{equation}
To solve this problem, we propose to solve the $n$ outputs one by one. When one output $y_i$ is considered, other outputs are fixed, $y_{i'}|_{i' = i}$. When only $y_i$ is considered, the problem in (\ref{equ:localproblem5}) is reduced to

\begin{equation}
\label{equ:localproblem6}
\begin{aligned}
\min_{y_i}~ &\sum_{i':\bfx_i \in \mathcal{N}_{i'}}\left \{ \frac{1}{k} \left [\Delta(y_i, z^*_{i',i}) - \bfw_{i'}^\top \Phi(\bfx_i,y_i) \right ] \right \}\\
s.t.~&
y_i = \overline{y}_i,if~(\bfx_i, \overline{y}_i)\in \mathcal{L}.
\end{aligned}
\end{equation}
To solve this problem, we discuss it in two cases.

\begin{itemize}
\item \textbf{Case I, $(\bfx_i, \overline{y}_i)\in \mathcal{L}$}: In this case, according to the constraint of (\ref{equ:localproblem6}), we assign $y_i = \overline{y}_i$,

\begin{equation}
\label{equ:y1}
\begin{aligned}
y_i = \overline{y}_i,if~(\bfx_i, \overline{y}_i)\in \mathcal{L}.
\end{aligned}
\end{equation}

\item \textbf{Case II, $\bfx_i \in \mathcal{U}$}: In this case, we obtain the follow optimal output as follows,

\begin{equation}
\label{equ:localproblem6}
\begin{aligned}
y_i = \underset{y\in \mathcal{Y}}{\arg\min} ~ &\sum_{i':\bfx_i \in \mathcal{N}_{i'}}\left \{ \frac{1}{k} \left [\Delta(y, z^*_{i',i}) - \bfw_{i'}^\top \Phi(\bfx_i,y) \right ] \right \},~if~\bfx_i\in \mathcal{U}.
\end{aligned}
\end{equation}
This is a minimization problem, and the objective function is a combination of functions over several neighborhoods which $\bfx_i$ belongs to. Each function is composed of two terms, one of them is a loss function between a candidate and a upper bound parameter. The other term is a negative function of the local predictor.

\end{itemize}

\section{Algorithms}
\label{sec:algorithm}

\subsection{Iterative learning algorithm}

Based on the optimization problem, we develop an iterative algorithm to learn the local structured output predictor and the outputs jointly. In each iteration, we first fix the both the local predictor parameters and the outputs to update the upper bound parameters, then update the local predictor parameters by fixing the outputs and the upper bound parameter, and finally fix local predictor parameters and the upper bound parameters to update the outputs. The iterations are repeated for $T$ times. The developed iterative algorithm is given in Algorithm 1.

\begin{itemize}
\item \textbf{Algorithm 1}. Iterative training algorithm of semi-supervised learning of local structured output predictor.
\item \textbf{Inputs}: Training set, $\mathcal{X}$.
\item {\textbf{Inputs}: Maximum iteration number, $T$}.

\item \textbf{Initialize} $(\bfw_i,y_i)|_{i=1}^n$;

\item \textbf{For $t=1,\cdots,T$}

\begin{itemize}
\item Update the upper bound parameters

\item  For $i=1,\cdots,n$

\begin{itemize}
\item For $j:\bfx_j\in \mathcal{N}_i$

\begin{equation}
\begin{aligned}
&z^*_{i,j} \leftarrow \underset{y_j'\in \mathcal{Y}}{\arg\max}
\left [\bfw_i^\top \left ( \Phi(\bfx_j,y'_j) - \Phi(\bfx_j,y_j) \right ) + \Delta(y_j, y_j')\right ],
\end{aligned}
\end{equation}

\end{itemize}

\item Update the local predictor parameters

\item For $i=1,\cdots, n$

\begin{equation}
\label{equ:updatew2}
\begin{aligned}
\bfw_i
\leftarrow
(1-\eta C)\bfw_i +  \frac{\eta}{k} \sum_{j:\bfx_j\in \mathcal{N}_i}
\left [ \left (  \Phi(\bfx_j,y_j)- \Phi(\bfx_j,z^*_{i,j}) \right ) \right ].
\end{aligned}
\end{equation}

\item Update the structured outputs

\item For $i = 1,\cdots,n$

\begin{itemize}
\item If $(\bfx_i,\overline{y}_i)\in \mathcal{L}$

\begin{equation}
\label{equ:outputcase1}
\begin{aligned}
y_i\leftarrow \overline{y}_i;
\end{aligned}
\end{equation}

\item Else

\begin{equation}
\label{equ:output}
\begin{aligned}
y_i \leftarrow \underset{y\in \mathcal{Y}}{\arg\min} ~ &\sum_{i':\bfx_i \in \mathcal{N}_{i'}}\left \{ \frac{1}{k} \left [\Delta(y, z^*_{i',i}) - \bfw_{i'}^\top \Phi(\bfx_i,y) \right ] \right \};
\end{aligned}
\end{equation}

\end{itemize}

\end{itemize}

\item \textbf{Output}: $\bfw_i|_{i=1}^n$.

\end{itemize}

\subsection{Algorithm of predicting structured output of test data point}

We have discussed how to learn local predictors from a training set. In this section, we discuss how to use these local predictors for the task of predicting structured output of a new-coming test data point. Suppose the input feature vector of the new-coming data point is $\bfx$, to predict its output, we first find which neighborhoods it belongs to. To this end, we first find its $k$ nearest neighbors from the training set, and denote the set of its  $k$ nearest neighbors as $\mathcal{N}_\bfx$. We assume $\bfx$ are in the neighbors of the data points in $\mathcal{N}_\bfx$, and use their local predictors to predict the structured outputs of $\bfx$. Given a candidate output, $y$, we use these local predictors to match it to $\bfx$ to obtain $k$ matching scores. The average matching score is used as the final matching score to match $\bfx$ and $y$. The candidate output which gives the maximum matching score is choose as the final structured output of $\bfx$,

\begin{equation}
\label{equ:predict}
\begin{aligned}
y^* = \underset{y\in \mathcal{Y}}{\arg\max} \frac{1}{k} \sum_{i:\bfx_i\in \mathcal{N}_\bfx} \bfw_i^\top \Phi(\bfx,y).
\end{aligned}
\end{equation}

The prediction algorithm for new-coming data point is given in Algorithm 2.

\begin{itemize}
\item \textbf{Algorithm 2}: Predicting the structured output of a new-coming test data point.

\item \textbf{Input}: $\bfx$ and $\mathcal{X}$;

\item Find the $k$-th nearest neighbor set of $\bfx$ from $\mathcal{X}$, $\mathcal{N}_\bfx$;

\item \textbf{Initialize} maximum matching score $s^* = -\infty$

\item \textbf{For $y\in \mathcal{Y}$}

\begin{equation}
\label{equ:predict}
\begin{aligned}
s_y =  \frac{1}{k} \sum_{i:\bfx_i\in \mathcal{N}_\bfx} \bfw_i^\top \Phi(\bfx,y);
\end{aligned}
\end{equation}

\begin{itemize}
\item If $s_y\geq s^*$

\begin{equation}
\begin{aligned}
s^* = s_y, y^* = y;
\end{aligned}
\end{equation}

\end{itemize}

\item \textbf{Output}: $y^*$

\end{itemize}

\section{Experiments}
\label{sec:Exp}

In this section, we evaluate the performance of the proposed semi-supervised structured output prediction method. The experiments are conducted on three benchmark data sets. We first compare it to several state-of-the-art semi-supervised structured output prediction method, and then analyze its sensitivity to parameters, $k$ and $C$, experimentally.

\subsection{Benchmark data sets}

We used three benchmark data sets in our experiments, which are discussed as follows.

\begin{itemize}

\item \textbf{Data set I - DDSM mammography image data set}: The last data set is a medical imaging data set which contains Mammography images \cite{heath2000digital}. This data set contains 2620 pairs of images and they belong to three different classes, normal, cancer, and benign. Thus this is a three class classification problem. Some example images of this data set are given in Fig. \ref{fig:DDSM}. To present the class of each image pair, we code it as a three-demesnial binary vector. For the $i$-th image pair, its structured output is a vector is $y_i = [y_{i1}, y_{i2}, y_{i3}]^\top$, where

\begin{equation}
\begin{aligned}
y_{i1} = \left\{\begin{matrix}
1, &if~the~i-th~image~pair~belongs~to~\textbf{normal}~class, \\
0, &otherwise.
\end{matrix}\right.
\\
y_{i2} = \left\{\begin{matrix}
1, &if~the~i-th~image~pair~belongs~to~\textbf{cancer}~class, \\
0, &otherwise.
\end{matrix}\right.
\\
y_{i3} = \left\{\begin{matrix}
1, &if~the~i-th~image~pair~belongs~to~\textbf{benign}~class, \\
0, &otherwise.
\end{matrix}\right.
\end{aligned}
\end{equation}
To compare a given input and its corresponding prediction, $y$, and $y^*$, we use the 0-1 loss,

\begin{equation}
\begin{aligned}
\Delta(y, y^*)  = 1,~ if~ y \neq y^*, ~and~ 0~ otherwise.
\end{aligned}
\end{equation}
To construct the feature vector of the $i$-th image pair, we use the bag-of-feature method. The images are split into small image patches, the image patches are quantized to a dictionary, and the quantization histogram is used as the feature vector of the image pair, $\bfx_i$. Moreover, the joint representation of a input vector $\bfx$ and a output vector $y$ is their tensor product,

\begin{equation}
\begin{aligned}
\Phi(\bfx,y) = \bfx \otimes y.
\end{aligned}
\end{equation}

\begin{figure}
\centering
\includegraphics[height=\textheight]{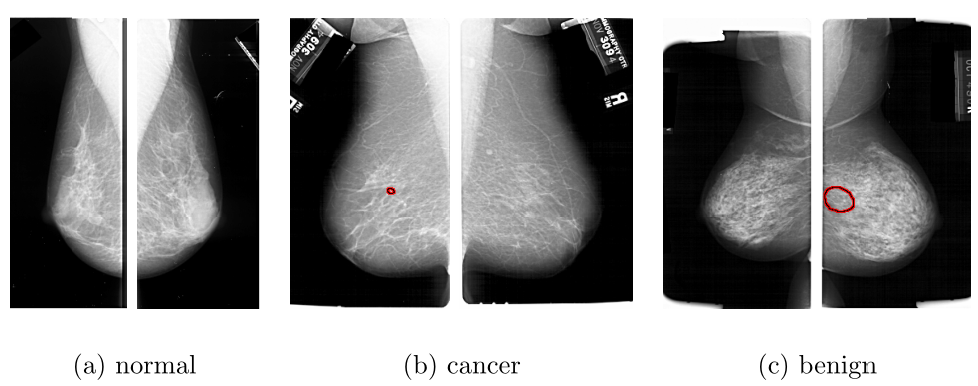}
  \caption{Example images of DDSM Mammography image data set.}\label{fig:DDSM}
\end{figure}

\item \textbf{Data set II - SUN natural image data set}: The second data set is a image data set \cite{xiao2010sun}. The class labels of the images of this data set is organized as a tree structure.  The tree has 15 different leaves. For each class, we randomly select 200 images from the data set to conduct our experiments, thus there are 3,000 images in our data set in total. Some example images of this data set are given in Fig. \ref{fig:SUN}. To extract features from each image, we calculate the HOG features. The structured output of a data point is a leave of the tree. We code it as a vector $\kappa$ dimension, $y = [y_1, \cdots, y_\kappa]^\top \in \{1,0\}^\kappa$, where $\kappa$ is the number of nodes of the tree. The $\iota$-th element of the vector, $y_\iota = 1$ if the $\iota$-th node is its predecessor or itself, and $0$ otherwise. Given a input $\bfx$ and a output $y$, the joint representation is

\begin{equation}
\begin{aligned}
\Phi(\bfx,y) = \bfx \otimes y.
\end{aligned}
\end{equation}
The loss function between $y$ and $y^*$, $\Delta(y,y^*)$, is defined as the height of their first common ancestor.

\begin{figure}[!tbh]
\centering
\includegraphics[width=0.23\textwidth]{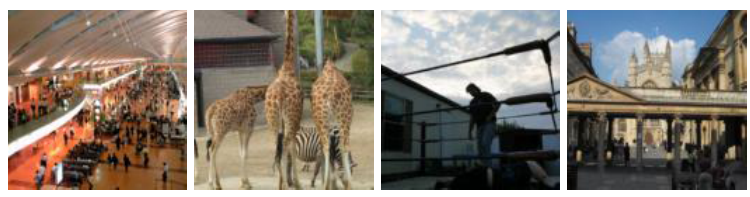}
\caption{Example images of SUN data set.}\label{fig:SUN}
\end{figure}

\item \textbf{Data set III - Cora research paper data set}: This data set is a set of computer science research papers \cite{mccallum2000efficient}. It contains 9,947 papers. To represent the papers, we extract two types of features. The first type is the frequencies of words of the papers, and the second type is the out-links of the citations of the papers. To extract the second type features, we remover the papers which do not have links to other papers, and a data set of 9,555 is used for the experiments. These papers belong to 7 different non-overlapping classes. Thus the structured output of each paper is coded as d 7-dimensional vector. The loss function defined for this structured output is the 0-1 loss. Moreover, the joint representation of a input vector $\bfx$ and a output vector $y$ is also their tensor product.

\item \textbf{Data set IV - Spanish news wire article sentence data set}: This data set is a data set for named entity recognition problem of natural language processing \cite{sang2002a}. It contains 300 sentences, and each sentence is used as a data point in our experiment. The length of each sentence is 9, and the corresponding output of a sentence, $y$,  is a sequence of labels of non-name and named entities. The joint representation $\Phi(\bfx,y)$ of a sentence, $\bfx$, and a sequence of labels, $y$, is defined as the histogram of state transition, and a set of emission features. The loss function to compare $y$ and $y^*$ is defined as a 0-1 loss, i.e., $\Delta(y,y^*) = 1$ if $y \neq y^*$, and $0$ otherwise.

\end{itemize}

\subsection{Experiment setup}

To conduct the experiments, we use the 10-fold cross validation strategy to split the training and test subsets. Given an entire data set, we randomly split it to ten subsets of equal sizes. Each subset is used as the test set, while the remaining nine subsets are combined and used as a training set. The training set are also randomly split to a labeled set and an unlabeled set. The Labeled set contains about $30\%$ of the training data points, and the unlabeled set contains the remaining $70\%$ data points. The proposed algorithm is performed to the training set to learn the local predictors, and then the local predictors are used to predict the structured outputs of the data points of the test set. The average structured loss over the test set is used to evaluate the performance of the proposed algorithm. Given a test set, $\mathcal{T}$ with $n_\mathcal{T}$ test data points, the average structured loss is defined as follows,

\begin{equation}
\begin{aligned}
Average ~Loss = \frac{1}{n_\mathcal{T}} \sum_{i:\bfx_i\in \mathcal{T}} \Delta(y_i, y_i^*)
\end{aligned}
\end{equation}
where $y_i$ is the true structured output of the $i$-th test data point, while $y_i^*$ is its corresponding predicted structured output.

\subsection{Comparison to state-of-the-arts}

We first compare the proposed algorithm to some state-of-the-art semi-supervised structured output prediction algorithms, including the algorithms proposed by Altun et al. \cite{altun2005maximum}, Brefeld and Scheffer \cite{brefeld2006semi}, Suzuki et al. \cite{suzuki2007semi}, and Jiang et al. \cite{jiang2015manifold}. All the four competing algorithm learn a single global predictor for all the data points, and our algorithm is the only algorithm that explores the local distribution of the data set, and learns the local predictors for different neighborhoods. Our algorithm is named as semi-supervised local structured output prediction algorithm (SSLSOP). The average losses of the compared algorithms over three different data sets are given in Table \ref{tab:compare}. From the results in Table \ref{tab:compare}, it is obvious that the proposed algorithm outperforms better than all competing algorithms significantly. For example, over data set I, only SSLSOP archives an average structured loss  lower than 0.400. The method of Jiang et al. \cite{jiang2015manifold} is the second best method. It is slightly better than the other mothers. This method also try to explore the local structure of the data set, and use it to regularize the learning of the structured output predictors. However, it still learns a global predictor, thus it cannot compete with the proposed method, SSLSOP. The results of the other algorithms are comparable to each other, and they are inferior to the SSLSOP. In summary, the results clearly show that multiple local predictors work better than one single predictor in the problems of structured output prediction.

\begin{table}
\centering
\caption{Average structured losses of the compared algorithms over three data sets.}
\label{tab:compare}
\begin{tabular}{|l|c|c|c|c|}
\hline
Method & Data set I & Data set II & Data set III& Data set IV \\\hline\hline
SSLSOP & \textbf{0.385} & \textbf{0.628} & \textbf{0.308}  & \textbf{0.450} \\\hline
Jiang et al. \cite{jiang2015manifold}& 0.401 & 0.677 & 0.383& 0.492\\\hline
Altun et al. \cite{altun2005maximum} & 0.438 & 0.738 & 0.497& 0.504 \\\hline
Brefeld and Scheffer \cite{brefeld2006semi} & 0.489 & 0.762 & 0.527& 0.511 \\\hline
Suzuki et al. \cite{suzuki2007semi} & 0.507 & 0.754 & 0.512 & 0.574\\\hline
\end{tabular}
\end{table}

\subsection{Sensitivity to parameters}

\begin{figure}
\centering
\includegraphics[width=\textwidth]{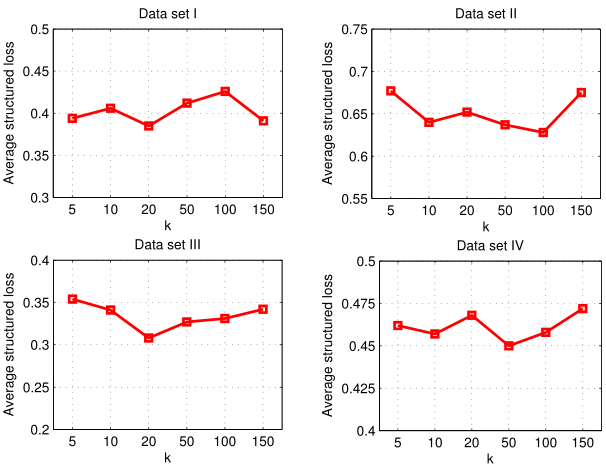}
\caption{Curves of sensitivity of SSLSOP to parameter $k$.}
\label{fig:k}
\end{figure}

\begin{figure}
\centering
\includegraphics[width=0.4\textwidth]{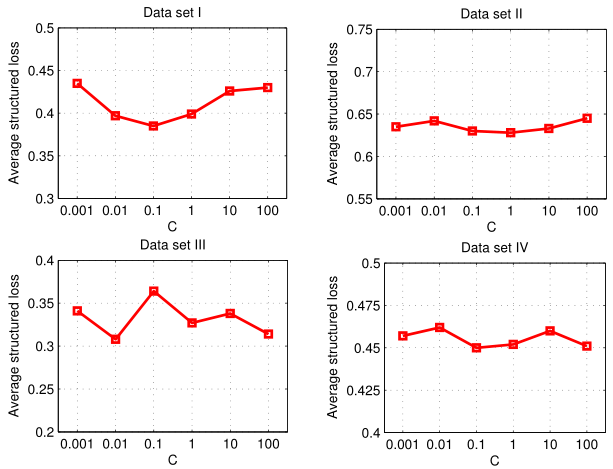}
\caption{Curves of sensitivity of SSLSOP to parameter $C$.}
\label{fig:C}
\end{figure}

In our objective, there are two important parameters, $k$, the size of each neighborhood, and $C$, the tradeoff parameter. To study the sensitivity of the algorithm to these two parameters, we plot the curves of the results with different values of the parameters. These curves of sensitivity to parameter $k$ over four benchmark data sets are shown in Fig. \ref{fig:k}. From the figures, we observe that the algorithm is stable to the changes of the parameter $k$. For different data sets, the optimal neighborhood sizes are different. For example, over the data set I, the lowest average structured loss is obtained at $k=20$, while for data set II, it is obtained at $k = 100$. The sensitivity curves of $C$ are given in Fig. \ref{fig:C}. From this figure, we also can see that the performances of the proposed algorithm are stable against the changes of the parameter $C$, especially over data set II. We cannot see a clear trend of the changes of the results corresponding to the changes of the values of $C$. This means that the algorithm is robust to the selection of parameter $C$.

\subsection{Repainting time analysis}

\begin{figure}
\centering
\includegraphics[width=0.8\textwidth]{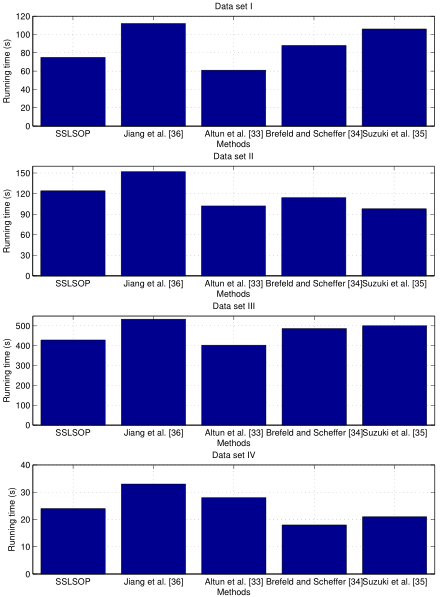}
\caption{Running time of the compared algorithms over the four data sets.}
\label{fig:Time}
\end{figure}

We are also interested in the running time of the proposed method, SSLSOP, and its competing methods. The running time of these methods over four benchmark data sets are given in Fig. \ref{fig:Time}. From the figure, we can see that the proposed method, SSLSOP, consumes the second shortest running time over three data sets. The only exception is the results over data set II. The least time consuming algorithm is the one proposed by Altun et al. \cite{altun2005maximum}, however, its prediction results are not satisfying. The most time consuming algorithm is the one proposed by Jiang et al. \cite{Jiang2015}, but its prediction results are not as good as SSLSOP.

\section{Conclusion}
\label{sec:Conclu}

In this paper, we investigate the problem of semi-supervised learning of structured output predictor. To handle the problem of diversity of the local distributions, we propose to learn local structured output predictors for neighborhoods of different data points. Moreover, we also propose to learn the missing outputs of the unlabeled data points. We build a new minimization problem to learn the local structured output predictors and the missing structured outputs simultaneously. This problem is modeled as the joint minimization of the local predictor complexity and the local structured output loss. The problem is optimized by gradient descent, and we designed an iterative algorithm to learn the local predictors. The experiments over benchmark data sets of medical image classification, natural image classification, computer science paper classification, and sentence part-of-speech tagging.

\section{{Future work}}
\label{sec:futurework}

{In the future, we will study how to fit the proposed algorithm to big data sets, by using big data processing framework, such as Map-Reduce of Hadoop software. When the data set is big, i.e., the number of data points, $n$, is large, we can the Hadoop distributed file system (HDFS) to store the data set. The entire data set is split into some small sub-sets, and different sub-sets are stored in different clusters. The proposed algorithm has three basic steps, and each of them can be parallelized easily. The three steps are listed as follows:
\begin{enumerate}
  \item finding the $k$ nearest neighbors of each data point,
  \item updating the local structured output predictor parameter of each neighborhood, and
  \item updating the outputs of each data point.
\end{enumerate}
To find the $k$ nearest neighbors of one data point from the distributed big data set, we can use the Map-Reduce framework. We use the Map function to find the $k$ nearest neighbors from each sub-set, and then use the Reduce function to find the final $k$ nearest neighbors from the results of the Map functions. To update the local structured output predictor parameter of one neighborhood, according to (\ref{equ:updatewi}), only the data points of the considered neighborhood is used. We can store the data points of each neighborhood in the same cluster, and use a Map function to update the parameter simultaneously for all the local structured output predictors. Similarly, the structured outputs also are calculated by exploring the data points of the neighborhoods, by using Map functions. }

\end{document}